\documentclass[runningheads]{llncs}
\usepackage{graphicx}
\usepackage{multirow}
\usepackage{algorithm, algorithmic}
\usepackage{subfigure}
\usepackage{amsfonts}
\usepackage{amsmath}
\usepackage{amssymb}
\usepackage{cite}
\usepackage{color}

\usepackage{makecell}
\usepackage{marvosym}

\usepackage{hyperref}
\hypersetup{hidelinks}
\hypersetup{
	colorlinks=true,
	linkcolor=blue,
	citecolor=blue
}

\begin{document}
\title{Unsupervised Domain Adaptation via Similarity-based Prototypes for Cross-Modality Segmentation}
\titlerunning{Similarity-based Prototypes for Cross-Modality Segmentation}
%
\author{Ziyu Ye, Chen Ju, Chaofan Ma, Xiaoyun Zhang\textsuperscript{(\Letter)}}
\authorrunning{Z. Ye et al.}
%
\institute{Cooperative Medianet Innovation Center,\\
	Shanghai Jiao Tong University, Shanghai, China \\
\email{\{ziyu\_ye,ju\_chen,chaofanma,xiaoyun.zhang\}@sjtu.edu.cn}}
\maketitle              

\begin{abstract}
Deep learning models have achieved great success on various vision challenges, but a well-trained model would face drastic performance degradation when applied to unseen data. Since the model is sensitive to domain shift, unsupervised domain adaptation attempts to reduce the domain gap and avoid costly annotation of unseen domains. This paper proposes a novel framework for cross-modality segmentation via similarity-based prototypes. In specific, we learn class-wise prototypes within an embedding space, then introduce a similarity constraint to make these prototypes representative for each semantic class while separable from different classes. Moreover, we use dictionaries to store prototypes extracted from different images, which prevents the class-missing problem and enables the contrastive learning of prototypes, and further improves performance. Extensive experiments show that our method achieves better results than other state-of-the-art methods.

\end{abstract}

\section{Introduction}
Medical image segmentation is a pixel-wise classification task, which is the basis of many clinical applications\cite{ref_sifa, ref_bsiris, ref_tis}. Though deep neural networks have made significant progress in medical image analysis\cite{ref_unet, ref_zongshu}, most supervised works have the assumption that enough annotated data is collected, which is prohibitively difficult in reality. In clinical scenarios, data collection is time-consuming and laborious, and pixel-wise annotations require expert knowledge of doctors. Hence, unsupervised domain adaptation (UDA) is introduced as an annotation-efficient method to help cross-modality medical image segmentation\cite{ref_yyx}.

UDA transfers the knowledge learned in a label-rich domain to a label-lacking domain, bridging the domain gap. Currently, there are two main streams for UDA. One is image-level adaptation\cite{ref_cycada, ref_2, ref_dsfn}, which aims to make the images of different domains appear similar, so that the label-lacking target domain can learn from the transferred source domain. The other stream focuses on feature-level adaptation\cite{ref_pnp, ref_adaoutput}, which aims to match the feature distributions with adversarial learning or contrastive learning. Besides, DualHierNet\cite{ref_yyx} also uses edges as self-supervision for the target domain, and EntMin\cite{ref_pointcloud} uses entropy minimization to narrow domain gaps.

\begin{figure}[t]
	\centering
	\includegraphics[width=0.85\textwidth]{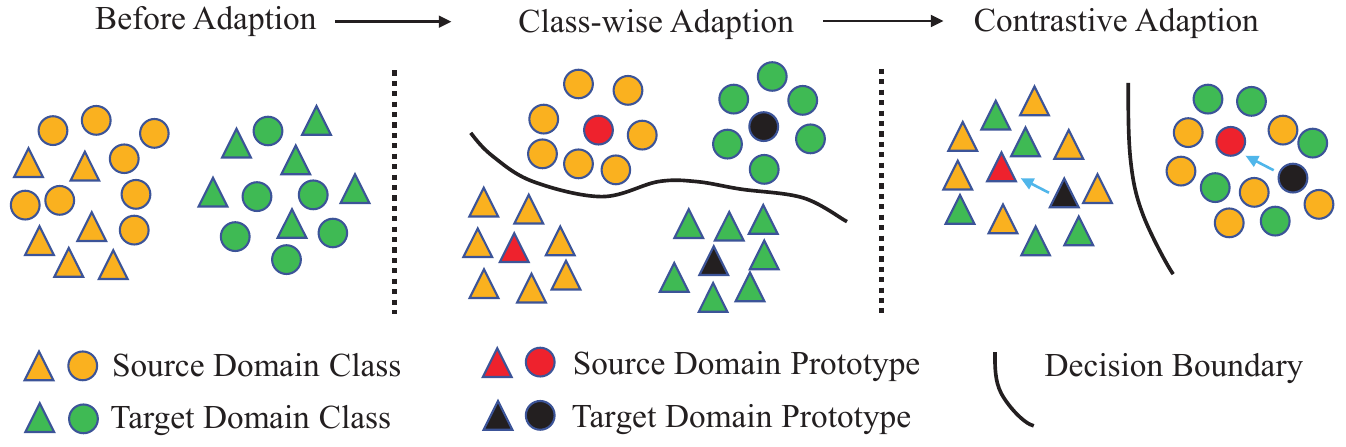}
	\caption{Illustration of our adaptation procedure. On the one hand, our method performs class-wise adaptation to align semantic features to their prototypes, on the other hand, we align class-wise prototypes across domains using the contrastive loss.}
	\label{fig:illustration}
\end{figure}

For methods based on image adaptation, most works only conduct the two-direction images translation between source and target domains separately, which may be insufficient to eliminate the domain gap. To this end, we propose a unified framework to fully exploit the two-direction translation results, and our network can be trained end-to-end. For methods based on feature adaptation, most works employ adversarial learning to make the semantic features indistinguishable to discriminators, which aligns features in an implicit way. In this paper, we explicitly align features to their prototypes using a class-wise similarity loss, which aims to minimize intra-class and maximize inter-class feature distribution difference. Then, with the help of feature dictionaries, we use the contrastive loss to align class-wise prototypes across domains, which further alleviates the domain shift problem. Fig.~\ref{fig:illustration} shows the illustration of our adaptation procedure.

\section{Methodology}
Given a labeled source dataset $\mathbb{D}_s=\left\{ x_{s}^{i},y_{s}^{i} \right\} _{i=1}^{N_s}$ and an unlabeled target dataset $\mathbb{D}_t=\left\{ x_{t}^{i} \right\} _{i=1}^{N_t}$, unsupervised domain adaptation (UDA) for semantic segmentation aims to train a model with supervision from $\mathbb{D}_s$ and information from $\mathbb{D}_t$ to narrow domain gap and improve segmentation performance on $\mathbb{D}_t$.

\subsection{Motivation}
In our method, we utilize class-wise feature prototypes to perform explicit feature alignment. Firstly, we use a similarity-based loss to regularize the embedded space, and the purpose is to boost feature consistency. Features of the same class are encouraged to be closer to the prototype, and prototypes of different classes are encouraged to be separable. Secondly, we use dictionary to store prototypes from various images, and then contrastive learning is used to improve feature adaptation across domains. We expect to adapt prototypes from target domain to source domain, so features of both domains are explicitly aligned.

\subsection{Proposed Framework}

\begin{figure*}[t]
	\centering
	\includegraphics[width=320pt]{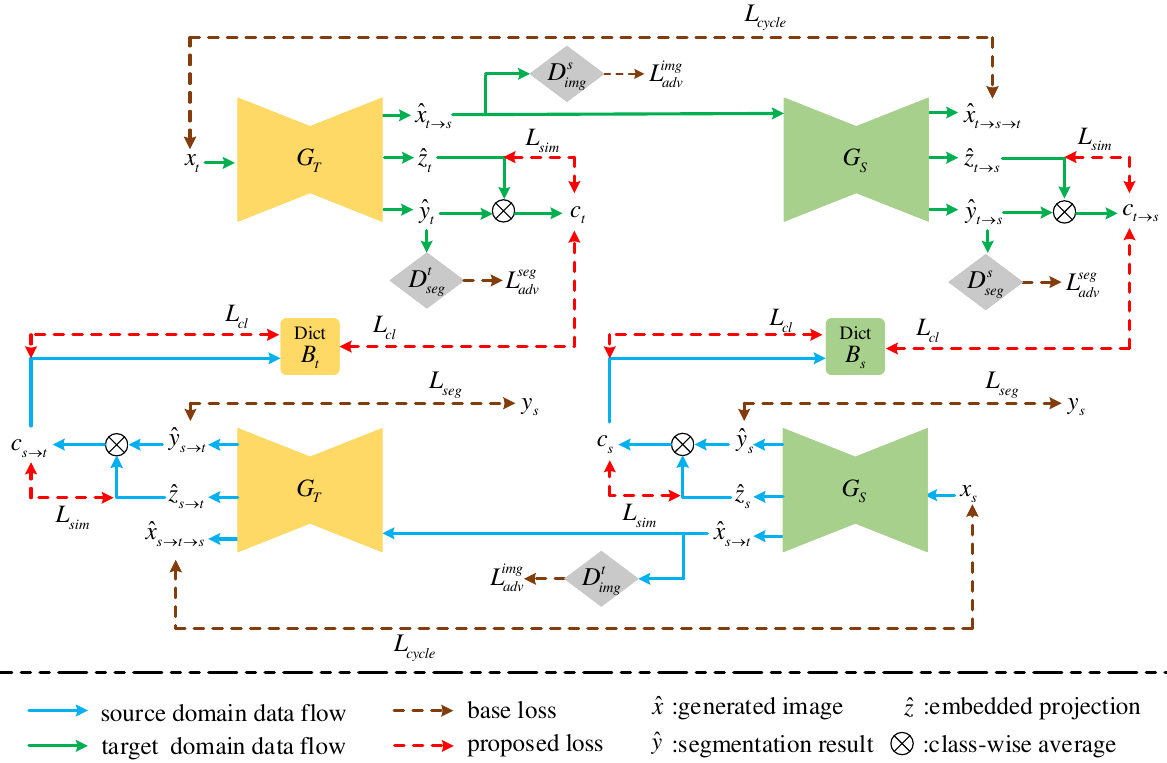}
	\caption{Our framework has a cycle structure, and mainly consists of $G_S$ and $G_T$. These modules have the same structure and output the translated image $\hat x$, segmentation result $\hat y$ and embedded projection $\hat z$. Prototypes $c$ are obtained by performing a class-wise average operation on $\hat z$ under supervision from $\hat y$. During training, only $c_s$ and $c_{s \rightarrow t}$ are stored into feature dictionaries $B_s$ and $B_t$. Besides the widely used circle consistency loss ${L_{cycle}}$, segmentation loss $L_{seg}$, adversarial loss $L_{adv}^{img}$ and $L_{adv}^{seg}$,  we additionally use the proposed loss $L_{sim}$ and $L_{cl}$ to perform explicit feature alignment.}
	\label{fig:framework}
\end{figure*}

\begin{figure}[h]
	\centering
	\includegraphics[width=320pt]{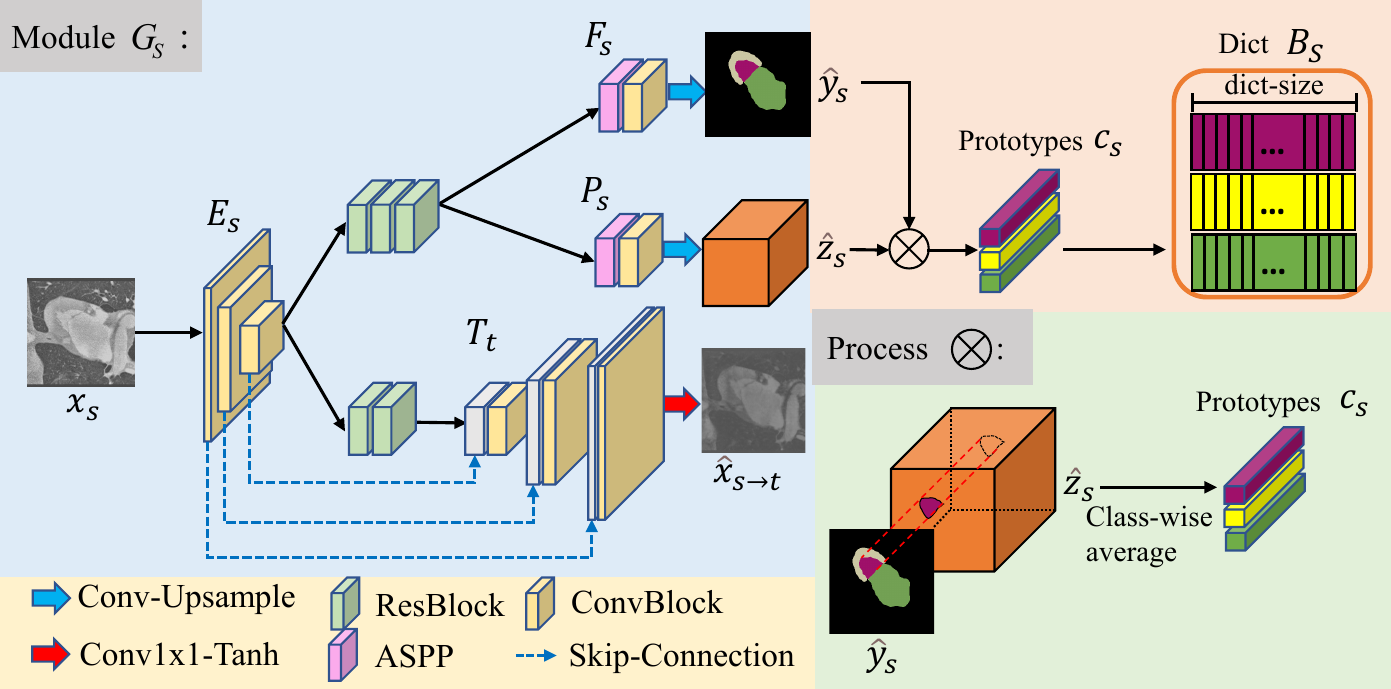}
	\caption{Details of $G_S$ and prototypes $c_s$. $G_S$ consists of a feature encoder $E_s$, an image generator $T_t$, and symmetric heads $F_s$ and $P_s$. These heads have same structures but different output channels. $F_s$ outputs the segmentation results ${\hat y_s}$ ,while $P_s$ outputs the embedded representation $\hat z_s$. The prototypes $c_s$ are obtained by performing class-wise average on $\hat z_s$ under supervision of ${\hat y_s}$. And we store these prototypes in Dictionary $B_s$.}
	\label{fig:detail}
\end{figure}

The overall framework is shown in Fig.~\ref{fig:framework}. It has a cycle structure inspired from Cycle-GAN\cite{ref_cyclegan}, and consists of two modules $G_S$ and $G_T$, which have the same structures, but process images of different domains. Concretely, $G_S$ processes source domains images, while $G_T$ processes target domain images. Structurally, we input an image for $G_S$ or $G_T$, and it will output the translated image $\hat x$, the segmentation result $\hat y$ and the embedded projection $\hat z$. During training, our framework is trained in a cycle manner. At each iteration, we calculate the class-wise prototypes $c$ from $\hat z$ under the supervision of $\hat y$. Note that we only store $c_s$ and $c_{s \rightarrow t}$ into feature dictionaries, since ${\hat y_s}$ and ${\hat y_{s \rightarrow t}}$ are trained under supervision of ground truth $y_s$, and we expect to adapt features from $c_{t \rightarrow s}$, $c_t$ to $c_s$, $c_{s \rightarrow t}$. During inference, we get the final output by directly averaging the target segmentation result $\hat y_t$ and the target-to-source segmentation result $\hat y_{t \rightarrow s}$.

To make it clear, we show the detailed structure of $G_S$ and the process to get prototypes $c_s$ in Fig.~\ref{fig:detail}. We introduce skip-connections to the image translation branch to help model convergence and image structure preservation\cite{ref_unet}. The segmentation head $C_s$ and projection head $P_s$ have the same structure but different output channels. This symmetrical design is proved effective for semantic feature extraction and regularization\cite{ref_auxhead}. We obtain prototypes $c_s$ from projection $z_s$ with the supervision from ${\hat y_s}$ by performing a class-wise average operation, and then we store $c_s$ into feature dictionary $B_s$.

In Fig.~\ref{fig:framework}, we denote the loss using the brown and red dash lines. Following GAN-based UDA methods\cite{ref_sasan, ref_sifa, ref_dsfn}, we use the cycle consistency loss $L_{cycle}$, segmentation loss $L_{seg}$ and adversarial loss $L_{adv}^{img}$, $L_{adv}^{seg}$ (see brown dash lines). Additionally, we design a class-wise similarity loss $L_{sim}$ to promote intra-class consistency and inter-class discrepancy at feature level. $L_{sim}$ is calculated between the projection $\hat z$ and the prototype $c$, which is calculated using $\hat z$ and $\hat y$. Besides, the contrastive loss $L_{cl}$ is used to align prototypes across domains, further reducing the domain gap and improving the model performance. The calculation of $L_{cl}$ needs the prototypes from feature dictionaries. At each iteration, prototypes $c_s$ and $c_{s \rightarrow t}$ are first used to calculate $L_{sim}$ and $L_{cl}$, then stored into Dict $B_s$ and $B_t$, respectively.

\subsection{Feature Prototypes and Class-wise Similarity Loss}
It is observed that the features of the same category tend to be clustered together\cite{ref_fsanchor}, but the features across different domains have significant discrepancies. To solve this issue, we regard class-wise prototypes as centers, and explicitly align features to their prototypes. As a result, prototypes of target domain are aligned to those of source domain.

\subsubsection{Feature Prototypes.}
Following\cite{ref_MPSCL, ref_auxhead}, we calculate the class-wise prototypes in a similar way, and the difference is that we use network segmentation result $\hat y$ instead of ground truth as supervision. As shown in Fig.~\ref{fig:detail}, we get prototypes $c_s$ by performing a class-wise average operation on $\hat z_s$ under supervision of $\hat y_s$. This procedure can be formulated as:
\begin{equation}
c_s^m = \frac{1}{{{N_m}}}\sum\limits_{i = 1}^{H \times W} \delta  \left( {{{\hat y}_s}[i],m} \right){\hat z_{s,i}},
\end{equation}
where $c_s^m$ denotes the prototype of the \textit{m}-th category from $x_s$, $N_m$ denotes the total pixels of the \textit{m}-th category, $\delta \left( {{{\hat y}_s}[i],m} \right)=1$ if the \textit{i}-th pixel of $\hat y_s$ belongs to category \textit{m}, and $\hat z_{s}$ is the embedded representation.

\subsubsection{Class-wise Similarity Loss.}
We propose a cosine similarity based loss $L_{sim}$ to explicitly regularize features in the embedding space, and we impose the constraint to $\hat z$. Taking $\hat z_s$ for example, the proposed loss $L_{sim}$ is the summation of the following $L_{sc}$ and $L_{dc}$.
\begin{align}
\label{eqn7} {L_{sc}} &= \frac{1}{C}\sum\limits_{m = 1}^C {\frac{1}{{{N_m}}}} \sum\limits_{i = 1}^{H \times W} \delta  ({{\hat y}_s}[i],m)\left( {1 - \cos {\mathop{\rm sim}\nolimits} \left( {c_s^m,{{\hat z}_{s,i}}} \right)} \right), \\
\label{eqn8} {L_{dc}}{\rm{ }} &= \frac{1}{{{N_c}}}\sum\limits_{m = 1}^C {\sum\limits_{n = m + 1}^C {(1 + \cos sim(c_s^m,c_s^n))} },
\end{align}
where $\cos sim(u,v) = \frac{{{u^T}v}}{{\left\| u \right\|\left\| v \right\|}}$ denotes the cosine similarity, $C$ denotes the number of categories in the image, ${N_c} = \frac{{C!}}{{2!(C - 2)!}}$ denotes the number of combinations. $L_{sc}$ becomes minimal when the similarity between category prototype $c_s^m$ and representation $\hat z_{s,i}$ is maximal. This aims to minimize the intra-class feature discrepancy. In $L_{dc}$, the similarity is calculated between prototypes of different classes, and it becomes minimal when these prototypes are as dissimilar as possible. This aims to maximize inter-class variance.

For target domain images, since the ${\hat y_t}$ and ${\hat y_{t \to s}}$ are trained without ground truth supervision, we use the pixel-wise predictions of high confidence to supervise $\hat z_t$ and $\hat z_{t \rightarrow s}$, and get the prototypes $c_t$ and $c_{t \rightarrow s}$. The similar idea has already been proven successful in pseudo-labeling\cite{ref_bdl}.

\subsection{Contrastive Loss via Feature Dictionaries}
To align features across domains and boost representative embedded projection, we use dictionaries to store class-wise prototypes from various images, which avoids the category missing problem and enables contrastive learning.

\subsubsection{Feature Dictionaries.}
In our framework, Dict $B_s$ and $B_t$ are used to store prototypes from $x_s$ and $x_{s \rightarrow t}$, respectively. Following \cite{ref_dict}, each dictionary has category labels as the keys and the values of each key are prototypes. We denote $B_s^c$ as the source domain dictionary accessed with category key $c$, and its shape is $[depth \times dict\_size]$. $B_s$ and $B_t$ are updated at every iteration, and old prototypes will be de-queued if the dictionary is full.

\subsubsection{Contrastive Loss of Prototypes.}
Taking $c_s^m$ (category $m$ from source image $x_s$) as an example, we first calculate the cosine similarity between $c_s^m$ and all prototypes stored in the dictionary $B_s$. Then for each category, we calculate the average of the highest \textit{k} similarity values. And the contrastive loss can be formulated as:
\begin{align}
&[v_1^{m,c},v_2^{m,c},...,v_L^{m,c}] = \cos sim(c_s^m,[d_{s,1}^c,d_{s,2}^c,...,d_{s,L}^c]), \\
&{v^{m,c}} = \frac{1}{k}\sum\limits_{i = 1}^k {topk(v_1^{m,c},v_2^{m,c},...,v_L^{m,c})}, \\
&{L_{cl}} =  - \frac{1}{C}\sum\limits_{m = 1}^C {\log } \frac{{\exp \left( {{v^{m,m}}/\tau } \right)}}{{\sum\limits_{i = 1,i \ne m}^C {\exp } \left( {{v^{m,i}}/\tau } \right) + \exp \left( {{v^{m,m}}/\tau } \right)}},
\end{align}
where $v_i^{m,c}$ denotes the cosine similarity between $c_s^m$ and $d_{s,i}^c$, $d_{s,i}^c$ denotes \textit{i}-th value from category $c$ of dictionary $B_s$, and $\tau$ is the temperature factor.

The contrastive loss not only makes the representation discriminative in embedding space, but also pulls target features closer to the source. Thus both domains are explicitly aligned at the feature level.

\subsubsection{Overall Objectives.}
Following~\cite{ref_sasan, ref_sifa, ref_dsfn}, the widely used cycle consistency loss $L_{cycle}$, segmentation loss $L_{seg}$ and adversarial loss $L_{adv}^{img}$, $L_{adv}^{seg}$ are also employed in our training process, we denote them as $L_{base}$. By adding our proposed similarity-base losses, and the overall objectives can be formulated as:
\begin{equation}
L_{all} = L_{base} + {\lambda _1}{L_{sim}} + {\lambda _2}{L_{cl}},
\end{equation}
where ${\lambda _1},{\lambda _2}$ are balance parameters.

\begin{figure}[t]
	\centering
	\includegraphics[width=\textwidth]{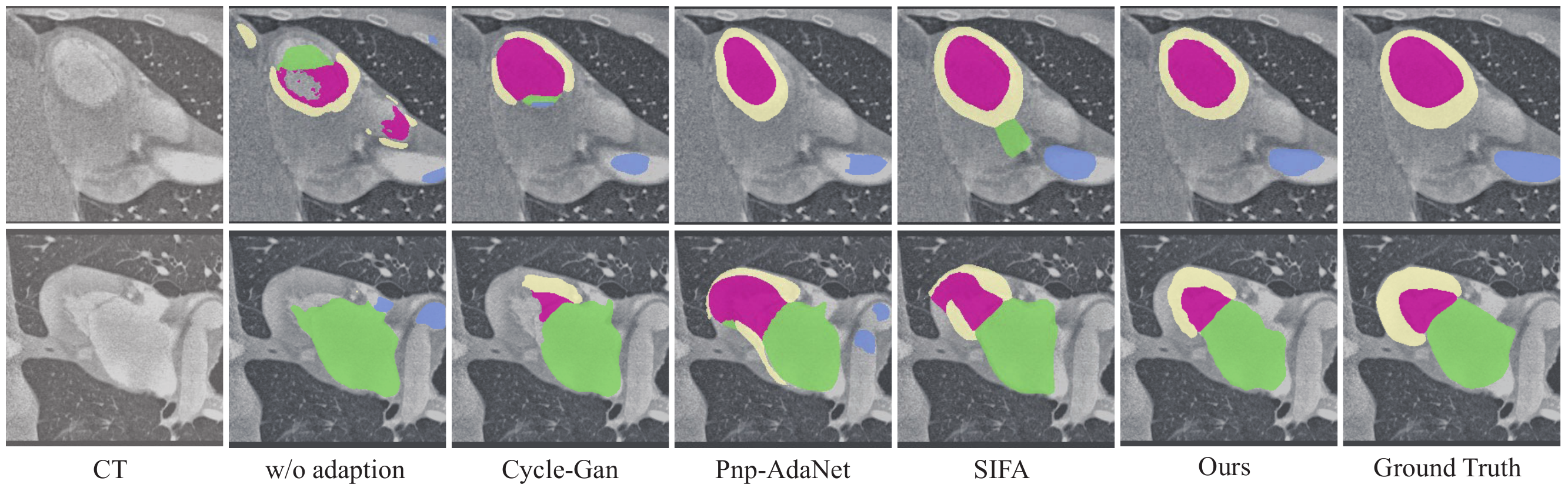}
	\caption{Visual comparison of representative methods. The structures of MYO, LAC, LVC, AA are denoted by \textcolor[rgb]{0.996,0.976,0.392}{yellow}, \textcolor[rgb]{0.565,0.792,0.412}{green}, \textcolor[rgb]{0.780,0.082,0.522}{red},
	\textcolor[rgb]{0.463,0.573,0.831}{blue} colors, respectively.}
	\label{fig:visual}
\end{figure}

\begin{table}[t]
	\centering
	\caption{Results of the MRI $\to$ CT task for four cardiac structures on \textit{MMWHS}.}
	\label{tab_compare}
	\begin{tabular}{lcccccccccc}
		\hline
		\multicolumn{1}{c|}{\multirow{2}{*}{Methods}} & \multicolumn{5}{c|}{Volumetric Dice$\uparrow$}                                                                                               & \multicolumn{5}{c}{Volumetric ASD$\downarrow$}                                                                          \\ \cline{2-11} 
		\multicolumn{1}{c|}{}                         & \multicolumn{1}{c|}{AA}   & \multicolumn{1}{c|}{LAC}  & \multicolumn{1}{c|}{LVC}  & \multicolumn{1}{c|}{MYO}  & \multicolumn{1}{c|}{Avg.} & \multicolumn{1}{c|}{AA}   & \multicolumn{1}{c|}{LAC}  & \multicolumn{1}{c|}{LVC}  & \multicolumn{1}{c|}{MYO}  & Avg. \\ \hline
		\multicolumn{1}{l|}{Supervised training}      & \multicolumn{1}{c|}{92.7} & \multicolumn{1}{c|}{91.1} & \multicolumn{1}{c|}{91.9} & \multicolumn{1}{c|}{87.7} & \multicolumn{1}{c|}{90.9}    & \multicolumn{1}{c|}{1.5}  & \multicolumn{1}{c|}{3.5}  & \multicolumn{1}{c|}{1.7}  & \multicolumn{1}{c|}{2.1}  & 2.2     \\
		\multicolumn{1}{l|}{W/o adaptation}           & \multicolumn{1}{c|}{28.4} & \multicolumn{1}{c|}{27.7} & \multicolumn{1}{c|}{4.0}  & \multicolumn{1}{c|}{8.7}  & \multicolumn{1}{c|}{17.2}    & \multicolumn{1}{c|}{20.6} & \multicolumn{1}{c|}{16.2} & \multicolumn{1}{c|}{-}    & \multicolumn{1}{c|}{48.4} & -       \\ \hline
		\multicolumn{1}{l|}{PnP-AdaNet\cite{ref_pnp}}       & \multicolumn{1}{c|}{74.0} & \multicolumn{1}{c|}{68.9} & \multicolumn{1}{c|}{61.9} & \multicolumn{1}{c|}{50.8} & \multicolumn{1}{c|}{63.9}    & \multicolumn{1}{c|}{12.8} & \multicolumn{1}{c|}{6.3}  & \multicolumn{1}{c|}{17.4} & \multicolumn{1}{c|}{14.7} & 12.8    \\
		\multicolumn{1}{l|}{SynSeg-Net\cite{ref_synseg}}    & \multicolumn{1}{c|}{71.6} & \multicolumn{1}{c|}{69.0} & \multicolumn{1}{c|}{51.6} & \multicolumn{1}{c|}{40.8} & \multicolumn{1}{c|}{58.2}    & \multicolumn{1}{c|}{11.7} & \multicolumn{1}{c|}{7.8}  & \multicolumn{1}{c|}{7.0}  & \multicolumn{1}{c|}{9.2}  & 8.9     \\
		\multicolumn{1}{l|}{AdaOutput\cite{ref_adaoutput}}  & \multicolumn{1}{c|}{65.2} & \multicolumn{1}{c|}{76.6} & \multicolumn{1}{c|}{54.4} & \multicolumn{1}{c|}{43.6} & \multicolumn{1}{c|}{59.9}    & \multicolumn{1}{c|}{17.9} & \multicolumn{1}{c|}{5.5}  & \multicolumn{1}{c|}{5.9}  & \multicolumn{1}{c|}{8.9}  & 9.6     \\
		\multicolumn{1}{l|}{CycleGAN\cite{ref_cyclegan}}    & \multicolumn{1}{c|}{73.8} & \multicolumn{1}{c|}{75.7} & \multicolumn{1}{c|}{52.3} & \multicolumn{1}{c|}{28.7} & \multicolumn{1}{c|}{57.6}    & \multicolumn{1}{c|}{11.5} & \multicolumn{1}{c|}{13.6} & \multicolumn{1}{c|}{9.2}  & \multicolumn{1}{c|}{8.8}  & 10.8    \\
		\multicolumn{1}{l|}{CyCADA\cite{ref_cycada}}        & \multicolumn{1}{c|}{72.9} & \multicolumn{1}{c|}{77.0} & \multicolumn{1}{c|}{62.4} & \multicolumn{1}{c|}{45.3} & \multicolumn{1}{c|}{64.4}    & \multicolumn{1}{c|}{9.6}  & \multicolumn{1}{c|}{8.0}  & \multicolumn{1}{c|}{9.6}  & \multicolumn{1}{c|}{10.5} & 9.4     \\
		\multicolumn{1}{l|}{EntMin\cite{ref_pointcloud}}        & \multicolumn{1}{c|}{83.0} & \multicolumn{1}{c|}{\textbf{81.3}} & \multicolumn{1}{c|}{67.2} & \multicolumn{1}{c|}{58.4} & \multicolumn{1}{c|}{72.5}    & \multicolumn{1}{c|}{\textbf{2.9}}  & \multicolumn{1}{c|}{\textbf{2.7}}  & \multicolumn{1}{c|}{6.3}  & \multicolumn{1}{c|}{6.4}  & \textbf{4.6}     \\
		\multicolumn{1}{l|}{SIFAv1\cite{ref_sifav1}}        & \multicolumn{1}{c|}{81.1} & \multicolumn{1}{c|}{76.4} & \multicolumn{1}{c|}{75.7} & \multicolumn{1}{c|}{58.7} & \multicolumn{1}{c|}{73.0}    & \multicolumn{1}{c|}{10.6} & \multicolumn{1}{c|}{7.4}  & \multicolumn{1}{c|}{6.7}  & \multicolumn{1}{c|}{7.8}  & 8.1     \\
		\multicolumn{1}{l|}{SIFAv2\cite{ref_sifa}}          & \multicolumn{1}{c|}{81.3} & \multicolumn{1}{c|}{79.5} & \multicolumn{1}{c|}{73.8} & \multicolumn{1}{c|}{61.6} & \multicolumn{1}{c|}{74.1}    & \multicolumn{1}{c|}{7.9}  & \multicolumn{1}{c|}{6.2}  & \multicolumn{1}{c|}{5.5}  & \multicolumn{1}{c|}{8.5}  & 7.0     \\
		\multicolumn{1}{l|}{DSFN\cite{ref_dsfn}}            & \multicolumn{1}{c|}{\textbf{84.7}} & \multicolumn{1}{c|}{76.9} & \multicolumn{1}{c|}{79.1} & \multicolumn{1}{c|}{62.4} & \multicolumn{1}{c|}{75.8}    & \multicolumn{1}{c|}{-}    & \multicolumn{1}{c|}{-}    & \multicolumn{1}{c|}{-}    & \multicolumn{1}{c|}{-}    & -       \\
		\multicolumn{1}{l|}{Ours}                     & \multicolumn{1}{c|}{82.6} & \multicolumn{1}{c|}{\textbf{81.3}} & \multicolumn{1}{c|}{\textbf{81.7}} & \multicolumn{1}{c|}{\textbf{64.3}} & \multicolumn{1}{c|}{\textbf{77.5}}    & \multicolumn{1}{c|}{6.1}  & \multicolumn{1}{c|}{6.0}  & \multicolumn{1}{c|}{\textbf{3.6}}  & \multicolumn{1}{c|}{\textbf{4.8}}  & 5.1     \\ \hline
	\end{tabular}
	\vspace{-10pt}
\end{table}

\begin{figure}[htbp]
    \centering
    \begin{minipage}[t]{0.48\textwidth}
    \centering
    \includegraphics[width=\textwidth]{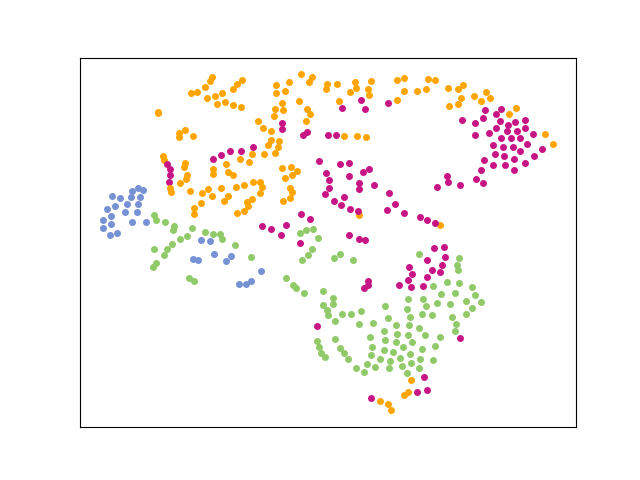}
    \end{minipage}
    \centering
    \begin{minipage}[t]{0.48\textwidth}
    \centering
    \includegraphics[width=\textwidth]{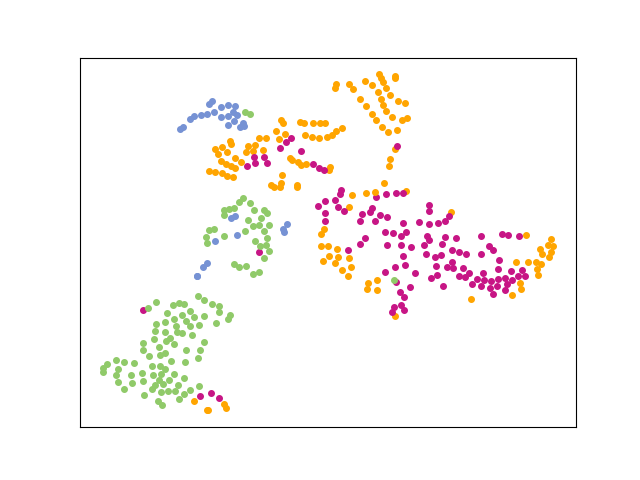}
    \end{minipage}
    \vspace{-20pt}
    \caption{t-SNE visualization of foreground features in Fig.\ref{fig:visual}. \textbf{Left:} The results without the proposed losses, \textbf{Right:} The results with the proposed losses.}
    \label{fig:tsne}
\end{figure}

\begin{table*}[t]
	\begin{minipage}{\textwidth}
		\centering
		\caption{Ablation study of the proposed losses}
		\label{tab_loss}
		\begin{tabular}{p{0.8cm}|c|c|c|c|c|c|c|c|c|c|c}
			\hline
			\multicolumn{2}{c|}{Components} & \multicolumn{5}{c|}{Dice$\uparrow$} & \multicolumn{5}{c}{ASD$\downarrow$} \\ \hline
			$L_{sim}$  & $L_{cl}$      & AA   & LAC  & LVC  & MYO  & Avg. & AA   & LAC  & LVC  & MYO  & Avg.  \\ \hline
			\makecell[c]{$\times$}        & $\times$      & 79.7 & 82.9 & 77.5 & 58.1 & 74.5    & 7.4  & \textbf{4.3}  & 4.1  & \textbf{4.3}  & \textbf{5.0}      \\
			\makecell[c]{$\checkmark$}    & $\times$      & 77.7 & \textbf{83.0} & 78.0 & 61.8 & 75.2    & 9.5  & 5.0  & 4.8  & 4.7  & 6.0      \\
			\makecell[c]{$\checkmark$}    & $\checkmark$  & \textbf{82.6} & 81.3 & \textbf{81.7} & \textbf{64.3} & \textbf{77.5}    & \textbf{6.1}  & 6.0  & \textbf{3.6}  & 4.8  & 5.1      \\ \hline
		\end{tabular}
	\end{minipage}
	\\[12pt]
	\begin{minipage}{0.5\textwidth}
		\centering
		\caption{Ablation study of three methods to utilize feature dictionaries}
		\label{tab_dicts}
		\begin{tabular}{l|c|c|c|c|c}
			\hline
			\multicolumn{1}{c|}{\multirow{2}{*}{Methods}} & \multicolumn{5}{c}{Dice$\uparrow$}                                                                                                                                \\ \cline{2-6} 
			\multicolumn{1}{c|}{}                         & AA                             & LAC                            & LVC                            & MYO                            & Avg.                           \\ \hline
			Max Similarity                                & 79.1                           & 84.1                           & 78.3                           & 61.2                           & 75.7                           \\
			Mean All                                      & 80.5                           & \textbf{82.1} & 79.1                           & 62.7                           & 76.1                           \\
			Mean Top-k                                    & \textbf{82.6} & 81.3                           & \textbf{81.7} & \textbf{64.3} & \textbf{77.5} \\ \hline
		\end{tabular}
	\end{minipage}
	\begin{minipage}{0.5\textwidth}
		\centering
		\caption{Ablation study of dict sizes $S$}
		\label{tab_dictp}
		\begin{tabular}{cl|c|c|c|c|c}
\hline
\multicolumn{2}{c|}{\multirow{2}{*}{$S$}} & \multicolumn{5}{c}{Dice$\uparrow$}                                                                                                                                \\ \cline{3-7} 
\multicolumn{2}{c|}{}                     & AA                             & LAC                            & LVC                            & MYO                            & Avg.                           \\ \hline
\multicolumn{2}{c|}{w/o}                 & 77.7                           & 83.0                           & 78.0                           & 61.8                           & 75.2                           \\ \hline
\multicolumn{2}{c|}{200}                 & 81.1                           & \textbf{84.8} & 81.6                           & 58.4                           & 76.5                           \\
\multicolumn{2}{c|}{400}                 & \textbf{82.6} & 81.3                           & \textbf{81.7} & \textbf{64.3} & \textbf{77.5} \\
\multicolumn{2}{c|}{600}                 & 80.3                           & 82.8                           & 80.3                           & 63.6                           & 76.8                           \\ \hline
\end{tabular}
	\end{minipage}
\end{table*}

\section{Experiments}
\subsection{Datasets and Details}
The proposed method is validated on the \textit{Multi-Modality Heart Segmentation Challenge 2017} (MMWHS) dataset\cite{ref_mmwhs}, which consists 20 unpaired MR and CT volumes data with their pixel-level ground truth of heart structures. The left ventricle blood cavity (LVC), the left atrium blood cavity (LAC), the myocardium of the left ventricle (MYO) and the ascending aorta (AA) are usually selected to evaluate the model segmentation performance. For a fair comparison, we use the preprocessed data released by \cite{ref_pnp, ref_sifa}, which contains randomly selected 16 volumes for training and 4 volumes for testing for both modalities. All data were first normalized to zero-mean and unit standard deviation, and then switched to [-1,1]. Each slice was cropped and resized to the size of 256$\times$256. These data were also augmented by rotation, scaling, and affine transformations.

\subsubsection{Implementations.}
The discriminators follow patchGAN\cite{ref_patchgan}, except that we replace log objective with least-squares loss for a stable training\cite{ref_lsgan}. We empirically set ${\lambda _1}=0.05$, ${\lambda _2}=0.02$, and the dictionary size $S$ is set to 400, temperature $\tau$ is set to 1, while top-\textit{k} is set to 20. Batch size and training epoch are set to 4 and 35, respectively. Besides, we use Adam optimizer\cite{ref_adam} with weight decay of $1 \times {10^{ - 4}}$, and the learning rate for discriminators is set to $2 \times {10^{ - 4}}$, while $3 \times {10^{ - 4}}$ for $G_S$ and $G_T$. To warm up training, we apply our proposed loss after the first epoch, and our model is trained on a NVIDIA Tesla V100 with PyTorch.

\subsection{Results and Analysis}
\subsubsection{Quantitative and Qualitative Analysis.}
Table \ref{tab_compare} shows the MRI$\to$CT adaptation performance comparison with other methods. Since our experiment is conducted under the same setting as \cite{ref_pnp, ref_sifa}, we directly refer to their paper results. As shown in Table.\ref{tab_compare}, the model without adaptation gets a poor performance on the unseen target domain. Methods based on image-alignment \cite{ref_cycada, ref_cyclegan, ref_synseg} and methods based on feature-alignment \cite{ref_pnp, ref_adaoutput, ref_pointcloud} can significantly improve the model results by narrowing the domain gap. \cite{ref_sifa, ref_dsfn} further improve the performance by taking both perspectives into account. Our proposed method outperforms these methods in terms of dice, and achieves an average result of 77.5\%, besides, we achieve an average ASD of 5.1, which is slightly worse than EntMin\cite{ref_pointcloud}. This indicates that our generated results may be not smooth on the boundary regions, while EntMin\cite{ref_pointcloud} conduct the entropy minimization to deal with high uncertainty of the boundary. Fig.~\ref{fig:visual} shows the visual comparison, and we choose \cite{ref_cyclegan, ref_pnp, ref_sifa} as the representative methods from different alignment perspectives. We visualize the feature distribution of Fig.\ref{fig:visual} using t-SNE\cite{ref_tsne} in Fig.\ref{fig:tsne}.

\vspace{-10pt}
\subsubsection{Ablation Study.}
Firstly, we evaluate the effectiveness of the proposed losses. As shown in Table.\ref{tab_loss}, when neither of the proposed losses were used, our method can be seen as a variant of \cite{ref_dsfn}, except that we redesign network structure to utilize low-level features to help image-translation and we do not use auxiliary task for feature adaptation. In this case, our method achieves an average Dice of 74.5\%. When only the class-wise similarity loss is used, the result gets a gain of 0.7\%. When both losses are applied, our model achieves an average dice of 77.5\%, surpassing other methods by a large margin.

Secondly, we test on several strategies to utilize prototypes in dictionaries. \textbf{Mean Top-k} means taking average of the top-\textit{k} similarity values. \textbf{Mean All} means taking average of all similarity values. \textbf{Max Similarity} means only use the largest similarity value. Table.\ref{tab_dicts} shows the results, and we can find that \textbf{Mean Top-k} achieves the best performance. This may due to the fact that sampling average can improve the robustness of similarity calculation, and the similarity will not be generalized to much.

Thirdly, we consider different dictionary sizes $S$. Table.\ref{tab_dictp} shows the results, we can achieve the best performance when $S$=$400$, which indicates that an appropriate dictionary size is necessary. A small dictionary may not have sufficient feature diversity, while a big dictionary may induce a slow updating of $L_{cl}$, as we calculate the average similarity using the top-\textit{k} strategy.

\vspace{-7pt}
\section{Conclusion}
This paper proposes a novel unsupervised domain adaptation framework for medical image segmentation. The framework is a unified network that can be trained end-to-end. We propose an innovative class-wise loss (calculated within a single sample) to boost feature consistency and learn representative prototype. Moreover, we conduct contrastive learning of prototypes (calculated with prototypes of multiple samples) to further improve feature adaptation across domains. Compared with existing adversarial learning based methods, we explicitly align features. Extensive experiments prove the effectiveness of our method, and show the superiority of the class-wise similarity loss and prototype contrastive learning via dictionary. In the future, we will test our method with different datasets and explore to apply it to domain generalization task.

%
%
\bibliographystyle{splncs04}
%

\end{document}